\documentclass[10pt,twocolumn,letterpaper]{article}

\usepackage{cvpr}              % To produce the CAMERA-READY version
\usepackage{enumitem}
\usepackage{dblfloatfix}
\usepackage{cuted} 
\usepackage{caption}
\usepackage[table]{xcolor}
\definecolor{cvprblue}{rgb}{0.21,0.49,0.74}
\usepackage[pagebackref=false,breaklinks,colorlinks,allcolors=cvprblue]{hyperref}
\usepackage[accsupp]{axessibility}
\makeatletter
\def\blfootnote{\xdef\@thefnmark{}\@footnotetext}
\makeatother

\def\paperID{3} % *** Enter the Paper ID here
\def\confName{CVPR}
\def\confYear{2026}
\newcommand{\formattedparagraph}[1]{\noindent \textbf{#1}}

\title{A Dataset and Evaluation for Complex 4D Markerless Human Motion Capture}

\author{\quad Yeeun Park$^{1, 2}$ \quad Miqdad Naduthodi$^{1, 2}$ \quad Suryansh Kumar$^{1, 2, 3, 4, *}$\\
Visual and Spatial AI Lab${^1}$, VCCM Section  \\ College of PVFA${^2}$, Department of ECEN$^3$, Department of CSCE$^4$, \\ Texas A\&M University, College Station, Texas, USA
}

\begin{document}
\maketitle

\vspace{-3mm}

\begin{strip}
\centering
\includegraphics[width=0.96\textwidth]{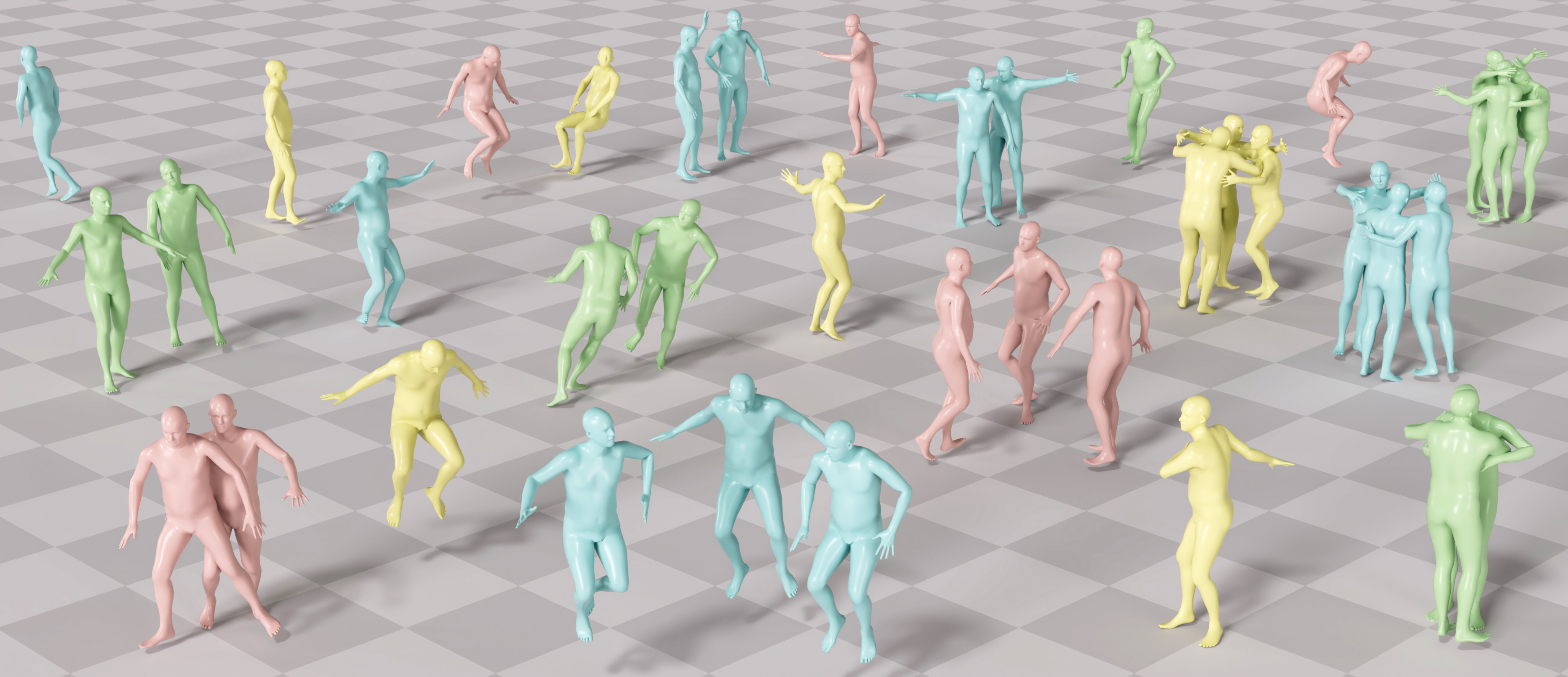}
\captionof{figure}{HUM4D provides multi-view RGB-D motion sequences with professional marker-based motion capture ground truth for complex multi-person interactions. The figure shows representative single and multi-person motions, highlighting challenging cases with severe occlusions and overlap. \textcolor{black}{\texttt{Project page:}} \url{https://parkyeeun23.github.io/HUM4D/}}
\label{fig:overview}
\vspace{-4mm}
\end{strip}
\begin{abstract}
Marker-based motion capture (MoCap) systems have long been the gold standard for accurate 4D human modeling, yet their reliance on specialized hardware and markers limits scalability and real-world deployment. Advancing reliable markerless 4D human motion capture requires datasets that reflect the complexity of real-world human interactions. Yet, existing benchmarks often lack realistic multi-person dynamics, severe occlusions, and challenging interaction patterns, leading to a persistent domain gap. In this work, we present a new dataset and evaluation for complex 4D markerless human motion capture. Our proposed MoCap dataset captures both single and multi-person scenarios with intricate motions, frequent inter-person occlusions, rapid position exchanges between similarly dressed subjects, and varying subject distances. It includes synchronized multi-view RGB and depth sequences, accurate camera calibration, ground-truth 3D motion capture from a Vicon system, and corresponding SMPL/SMPL-X parameters. This setup ensures precise alignment between visual observations and motion ground truth. Benchmarking state-of-the-art (SOTA) markerless MoCap models reveals substantial performance degradation under these realistic conditions, highlighting limitations of current approaches. We further examine the challenges posed by current SOTA baselines in handling occlusions, temporal coherence, etc. Our evaluation exposes critical gaps in SOTA models and identifies the potential of using multi-view data in model training to advance robust markerless 4D human MoCap\blfootnote{*Corresponding Author: Suryansh Kumar}.

\end{abstract}    
\section{Introduction}\label{sec:intro}

Marker-based human motion capture (MoCap) system has achieved great success in 4D human modeling \cite{Joo_2017_TPAMI, Simon_2017_CVPR, Joo_2015_ICCV, kumar2022organic, kumar2017spatio, kumar2020non}. Moving forward, the goal is to achieve similar, if not better, results with Markerless approaches \cite{shin2024wham, wang2025prompthmr}. As a result, Markerless human MoCap from images has witnessed remarkable progress over the past decade \cite{sigal2010humaneva, chatzitofis2020human4d, 6682899, wang2025prompthmr, shin2024wham, Mahmoodetal2019, Meshcapade2024}. Advances in deep neural network architectures combined with the availability of large-scale datasets with 3D ground truth have enabled increasingly accurate reconstruction of human pose and shape from monocular and multi-view imagery \cite{pavllo20193d, 6682899}. On benchmark datasets, current state-of-the-art (SOTA) models report steadily decreasing reconstruction errors, suggesting rapid maturation of deep learning-based approaches \cite{goel2023humans, Hao_2025_ICCV}.

Nevertheless, high performance on benchmark datasets does not  translate to robustness in real-world videos. A closer examination reveals that benchmark datasets impose structural constraints such as limited clothing variability, controlled indoor environments, moderate motion dynamics, restricted levels of occlusion, and predominantly single-person capture \cite{Bogo:CVPR:2014, Kocabas_PARE_2021, chatzitofis2020human4d, 6682899, Joo_2017_TPAMI}. As a result, trained models frequently fail when confronted with realistic multi-person interactions, rapid identity exchanges, severe inter-person occlusions, and dynamic scene complexity. Our preliminary experiments confirm that the SOTA models trained on popular benchmarks exhibit significant performance degradation when evaluated under realistic conditions.

These observations suggest that widely adopted datasets such as Human3.6M \cite{6682899}, CMU Panoptic \cite{Joo_2017_TPAMI}, and HUMAN4D \cite{chatzitofis2020human4d} are approaching saturation in terms of the complexity they offer. Despite being instrumental in advancing MoCap, these datasets capture controlled environments and basic motion-capture protocols and setups. So, given the current demand for large-scale Markerless Mocap models, a domain gap persists between benchmark performance and deployment in unconstrained, real-world environments. Addressing this gap requires not only architectural innovation, but also a fundamental rethinking of innovation and the augmentation of new datasets in the field that are timely and meet current demand.

% Constructing such a dataset is non-trivial. This is mainly due to a) required precision in ground-truth data using Multi-view synchronization across RGB and RGB-D sensors, b) Accuracy in geometric calibration is essential, c) Professional marker-based MoCap systems must be tightly aligned with visual observations to provide reliable ground truth. 

Meanwhile, acquisition of such a dataset is non-trivial due to \textbf{1)} the required precision in ground-truth data using Multi-view synchronization across RGB and RGB-D sensors, \textbf{2)} challenges with geometric calibration to meet an essential accuracy level, and \textbf{3)} the requirement of tight alignment of visual observation with professional marker-based MoCap systems to perform reliable ground-truth acquisition. In addition, capturing realistic interactions, including more than three participants, rapid motion transitions, occlusions, and object interactions introduces substantial logistical and technical challenges. These factors partly explain why existing datasets often avoid such complexity.

To bridge the existing gaps, we present HUM4D, a new dataset and evaluation for complex 4D markerless human MoCap in this paper. Our dataset comprises synchronized multi-view RGB and RGB-D sequences, precise camera calibration, and high-accuracy marker-based ground truth acquired using a professional MoCap system. We provide temporally aligned 3D shapes and pose trajectories, along with SMPL \cite{loper2015smpl} and SMPL-X \cite{SMPL-X:2019} parameters to facilitate research in parametric human modeling. The captured scenarios include single-person motions and multi-person interactions involving rapid position exchanges, dynamic occlusions, furniture interactions, and varying inter-subject distances (see Fig.\ref{fig:overview}). Importantly, the dataset emphasizes real-world variability rather than studio-staged setup.

%significantly improves the generalization performance of the existing models, validating its effectiveness as an excellent training resource and its importance for future markerless MoCap model development
%of popular approaches in this field

Beyond proposing a new MoCap dataset, we explored the  method's benchmarking and evaluation. To that end, we evaluated both popular and contemporary SOTA markerless MoCap methods on our dataset. The results reveal substantial performance degradation, highlighting the limitations of current models under realistic motion and scene complexity. We further observe that our dataset provides challenging real-world cases that could greatly enhance the generalization performance of existing models, and thus can serve as an excellent training resource for studying markerless MoCap model development. In this paper, we specifically aim to cover the following aspects:

\begin{itemize}[noitemsep, leftmargin=*]
    \item \textbf{A new dataset for complex human motion modeling}. We introduce a diverse collection of human shape, poses, and activities encompassing single-person dynamics and multi-person interactions, including fast motions, occlusions, identity exchanges, and object interactions.
    \item \textbf{Synchronized camera capture with ground truth}. The proposed dataset provides synchronized multi-view RGB and RGB-D images, precise camera calibration, and marker-based MoCap ground truth data. We additionally provide aligned SMPL and SMPL-X parameters for parametric human modeling and 4D reconstruction study.
    \item \textbf{Comprehensive evaluation and benchmarking}. We benchmark popular and current SOTA markerless MoCap methods on our dataset. Our evaluation shows significant performance gaps under realistic conditions, establishing a new, challenging testbed for the community.
\end{itemize}

%and demonstrates measurable gains through dataset-driven adaptation
\section{Related Work}
\label{sec:relatedwork}

\begin{table*}[t]
\centering
\caption{Comparison of our dataset to other benchmark datasets}
\label{table:table01}
\resizebox{\textwidth}{!}{%
\begin{tabular}{lccccccccc}
\hline
\textbf{Dataset} & \textbf{Multiperson} & \textbf{\# Subjects} & \textbf{Sparse or Dense} & 
\textbf{RGB} & \textbf{Depth} & \textbf{Annotations} & \textbf{Frame} & 
\textbf{Ground Truthing} \\
\hline
        MPII \cite{andriluka20142d} & No & 1 Person & Dense & Yes & No & 2D Pose, Activity Labels & 24k & Manual \\
        COCO \cite{lin2014microsoft} & No & More than 2 People & Dense & No & No & 2D Pose & 104k & Manual \\
        MOYO \cite{tripathi20233d} & No & 1 Person & Sparse & Yes & No & 3D Pose, Activity Labels, SMPL & $\approx$ 1.75M & Activity Labels \\
        UMPM \cite{van2011umpm} & Yes & More than 2 People & Sparse & Yes & No & 3D Pose, Activity Labels, SMPL & 400k & Activity Labels \\
        CMU Kitchen \cite{de2009guide} & No & 1 Person & Sparse & Yes & No & 3D Pose, Activity Labels & - & Activity Labels \\
        HumanEva \cite{sigal2010humaneva} & No & 1 Person & Both & Yes & No & 3D Pose, MoCap data & $\approx$ 80k & MoCap \\
        Human3.6M \cite{6682899} & No & 1 Person & Dense & Yes & Yes & 3D Pose, SMPL & 3.6M & MoCap \\
        CMU Panoptic \cite{Joo_2017_TPAMI} & Yes & More than 2 People & Dense & Yes & Yes & 3D Pose, SMPL & 154M & MoCap \\
        HuMMan \cite{cai2022humman} & Yes & 1 Person & Dense & Yes & Yes & 3D Pose, SMPL, Textured Mesh & 60M & Simulated \\
        HUMAN4D \cite{chatzitofis2020human4d} & Yes & Up to 2 People & Both & Yes & Yes & 3D Pose, MoCap, SMPL & 50K & MoCap \\
        \rowcolor{orange!40}
        Our Dataset & \textbf{Yes} & \textbf{Up to 3 People} & \textbf{Both} & \textbf{Yes} & \textbf{Yes} & \textbf{3D Pose, MoCap, SMPL} & \textbf{84k} & \textbf{MoCap} \\
        \hline
    \end{tabular}
    }
\end{table*}

In this section, we review prior datasets and discuss their implications in human mesh recovery and multi-person 3D reconstruction applications \cite{kumar2017spatio, kumar2016multi}.

\subsection{Datasets for 2D and 3D Human Motion Capture}
Early large-scale datasets such as MPII \cite{andriluka20142d} and COCO \cite{lin2014microsoft} were instrumental in advancing 2D pose estimation. Their diversity in appearance and scene context enabled robust keypoint detection in unconstrained imagery. However, the absence of depth maps or ground-truth 3D limits their applicability to full markerless MoCap and 4D modeling.

Subsequent datasets introduced accurate 3D supervision using professional marker-based systems. HumanEva \cite{sigal2010humaneva} and Human3.6M \cite{6682899} provide synchronized RGB images and accurate 3D pose annotations captured in controlled lab environments. These datasets have become standard benchmarks for 3D pose estimation. Yet, they predominantly feature single-person actions with limited clothing variability and minimal occlusion in studio conditions. Therefore, these datasets are not suitable for evaluating the performance of methods for multi-person interactions and related real-world complexities. To address this, 3DPW \cite{vonMarcard2018} captures outdoor in-the-wild human motion with SMPL-based annotations. Likewise, AMASS \cite{Mahmood:AMASS:2019} aggregates multiple MoCap datasets into a unified parametric representation, significantly expanding motion diversity for training human body models. While these datasets improve motion realism and coverage, they either lack synchronized multi-view RGB-D data or do not have dense multi-person interaction scenes with hardware-level synchronization.

CMU Panoptic dataset \cite{Joo_2017_TPAMI} significantly extended the MoCap scale by deploying a large multi-view camera dome to record rich social interactions involving multiple participants. Although it represents a milestone for multi-person capture, much of its 3D supervision relies on markerless keypoint triangulation and model-fitting pipelines rather than fully marker-based ground truth, which may introduce systematic inaccuracies in complex overlapping scenarios and lead to a loss of high-fidelity evaluations. HUMAN4D \cite{chatzitofis2020human4d} provides multi-view volumetric reconstructions and multi-person sequences, yet interactions remain relatively structured and typically involve only 2 actively interacting subjects in MoCap settings. More recent efforts, such as HuMMan \cite{cai2022humman}, incorporate dense RGB-D streams and parametric body annotations; however, portions of the supervision rely on synthetic augmentation or fitting-based estimation rather than direct marker-based capture.

\begin{figure}
    \centering
\includegraphics[width=\linewidth]{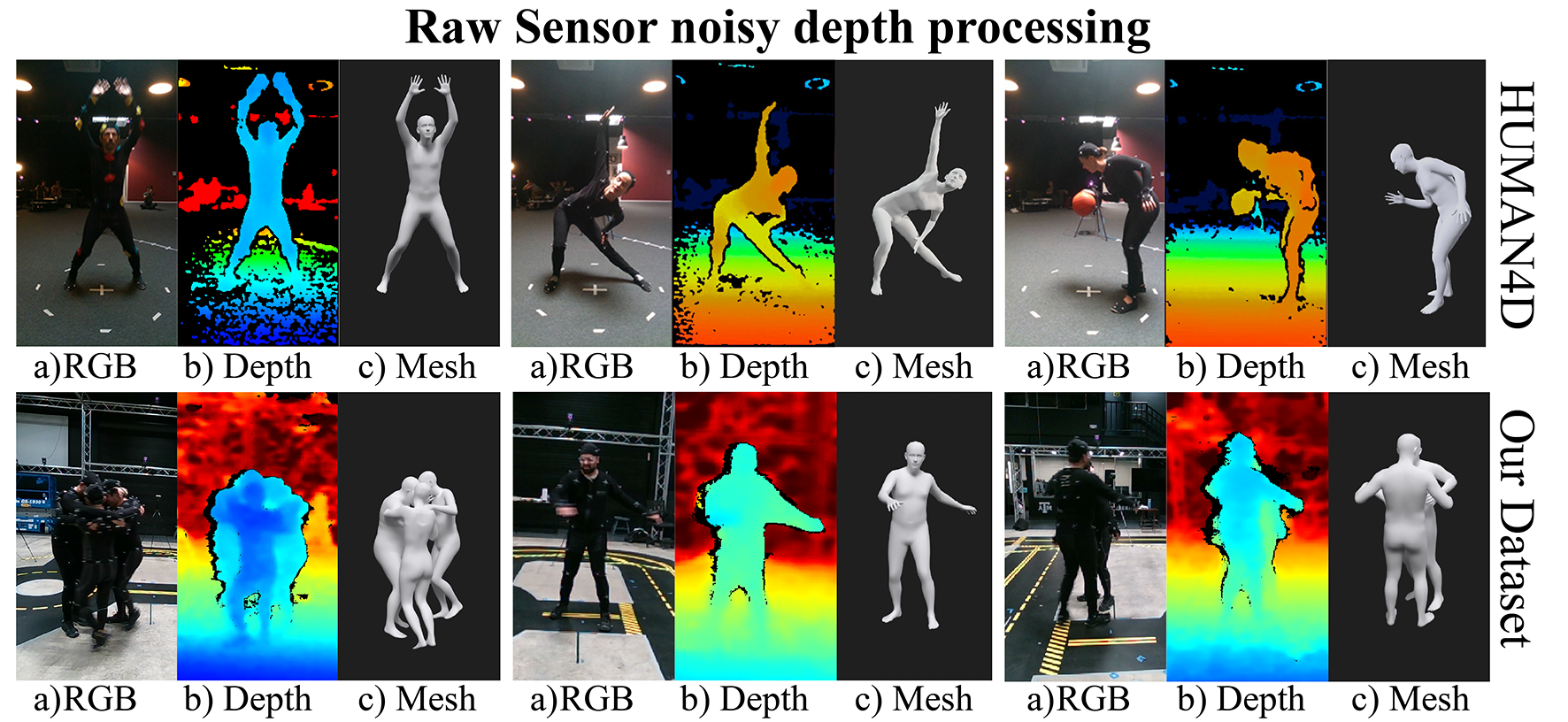}
\caption{Example noisy depth data acquisition from RGB-D sensor and our setup compared to recent HUMAN4D setup \cite{chatzitofis2020human4d}, showing raw RGB, depth image, and reconstructed SMPL mesh \cite{chatzitofis2020human4d} present in our dataset.}\label{fig:intro-fig}
\end{figure}

Overall, while existing datasets have substantially helped advance the field, none simultaneously provide: (i) synchronized multi-view RGB-D data, (ii) professional marker-based MoCap ground truth, and (iii) highly interactive scenarios involving more than two participants with complex, dynamic motion, all within a unified benchmark. \textit{This gap motivates the dataset introduced in this work}. A comparative summary regarding the same is provided in Table~\ref{table:table01}.

Recent advances in markerless human MoCap can broadly be categorized into three directions: (i) Human Mesh Recovery for reconstruction of parametric body model, (ii) articulated pose estimation, which focuses on recovering skeletal joint configurations, and (iii) full 3D human reconstruction, which aims to recover dense surface geometry over time. We review these lines of work below.

%\subsection{Human Mesh Recovery} 
% Human Mesh Recovery (HMR) methods estimate parametric body models such as SMPL~\cite{loper2023smpl} or SMPL-X from visual observations. These approaches typically rely on 2D/3D keypoints, multi-view images, or point clouds, and often use MoCap-derived supervision for training. For example, MoSh has been applied to Human3.6M to derive SMPL parameters, while multi-view keypoint triangulation in CMP Panoptic is used for parametric body fitting. 

% Despite strong performance on benchmarks, current HMR models are predominantly trained and evaluated on single-person or weakly interactive scenes. Severe inter-person occlusions, frequent body overlap, and extended volumetric coverage remain underrepresented in existing benchmarks. As a result, it remains unclear how well parametric reconstruction methods generalize to tightly coupled multi-person interactions. A dataset with multi-view RGB-D data and accurate ground truth in socially interactive settings is therefore essential for HMR research.

\smallskip
\formattedparagraph{\textit{(i)} Human Mesh Recovery (HMR)} aims to reconstruct parametric body models such as SMPL \cite{loper2015smpl} or SMPL-X \cite{SMPL-X:2019} from visual input. These approaches typically combine 2D or 3D keypoint supervision with parametric body fitting, either through direct regression or optimization-based refinement. For example, MoSh \cite{loper2014mosh} has been used to derive SMPL parameters from marker-based MoCap data in Human3.6M \cite{6682899}, while multi-view triangulated keypoints in CMU Panoptic \cite{Joo_2017_TPAMI} has supported parametric body fitting.

Despite strong performance on benchmark datasets, current HMR models are largely trained and evaluated on single-person or weakly interactive scenes. Severe inter-person occlusions, persistent body overlap, identity switching, and dense multi-person configurations remain underrepresented in training data. As a result, the generalization capacity of parametric reconstruction methods in tightly coupled social interactions remains unclear. A dataset providing synchronized multi-view RGB-D data together with accurate marker-based ground truth in complex interactive settings is therefore critical for advancing HMR research.

\smallskip
\formattedparagraph{\textit{(ii)} Multi-Person 3D Pose Estimation} aims to recover articulated skeletal configurations of multiple interacting individuals from monocular or multi-view observations. Recent approaches combine deep networks with model-based optimization to estimate 3D joint locations while enforcing anatomical constraints \cite{kolotouroslearning}. These methods show strong performance on widely used benchmarks such as Human3.6M \cite{6682899}, 3DPW~\cite{vonMarcard2018}, and CMU Panoptic \cite{Joo_2017_TPAMI}. Despite this progress, existing evaluations are largely conducted in controlled or moderately interactive environments where occlusion and body overlap remain limited. In realistic scenarios involving multiple people, articulated body parts frequently overlap or become partially invisible, leading to ambiguous 2D observations and unreliable keypoint detection. Such conditions significantly complicate the recovery of accurate 3D joint configurations. Furthermore, most existing datasets exhibit limited diversity in interaction patterns and motion dynamics, limiting the ability to systematically evaluate robustness to dense social interactions and rapid pose transitions. As a result, the generalization capability of current multi-person pose estimation methods in complex interaction scenarios remains underexplored.

% \subsection{3D Volumetric Reconstruction}

% 3D volumetric reconstruction aims to recover temporally consistent geometric representations of human motion from multi-view observations of human motion from multi-view observations~\cite{newcombe2015dynamicfusion}. Multi-view RGB-D data improve geometric completeness and surface fidelity, particularly in dynamic settings. However, accurate quantitive evaluation requires precisely synchronized ground truth motion capture.

% In multi-person scenarios, even minor temporal misalignment across RGB-D streams can introduce significant geometric artifacts and degrade reconstruction quality. Many existing datasets either lack hardware-level synchronization across sensors or do not provide professional marker-based ground truth for complex interactions. This limits the ability to systematically study markerlss multi-person volumetric reconstruction under realistic conditions.

\begin{figure}[t]
    \centering
\includegraphics[width=\linewidth]{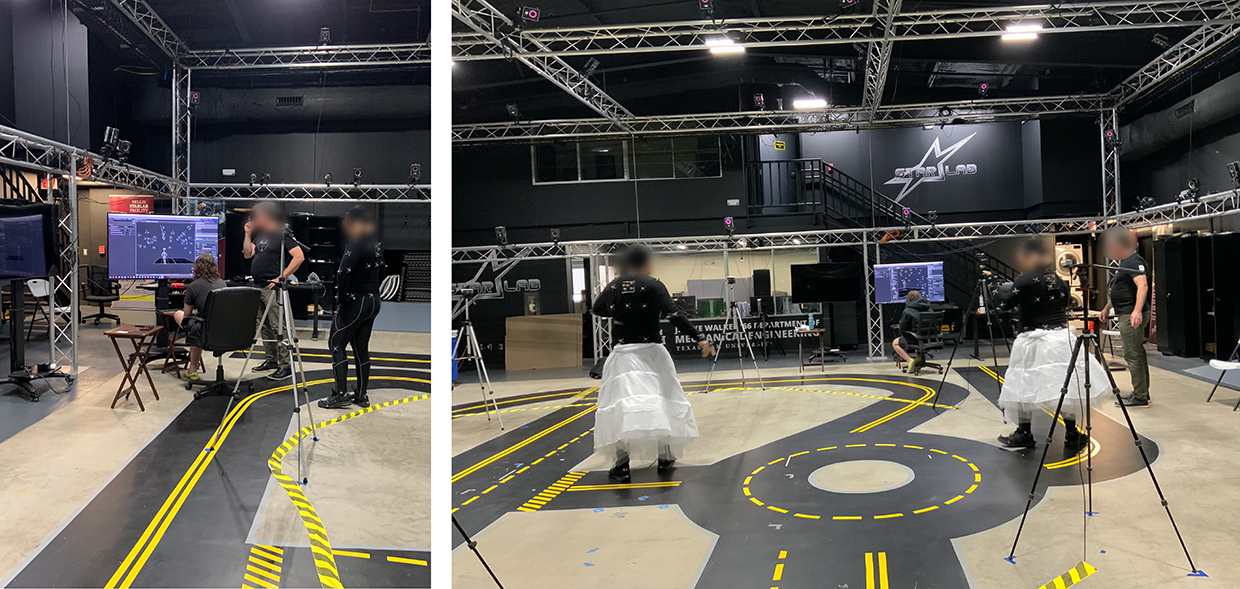}
\caption{\textbf{Capture Environment.} Professional MoCap studio for HUM4D data collection. The setup includes 44 synchronized infrared Vicon cameras for marker tracking and a multi-view RGB-D configuration for color and depth image sequence acquisition.}\label{fig:capture-stage}
\end{figure} 

%\subsection{3D Human Reconstruction}
\smallskip
\formattedparagraph{\textit{(iii)} 3D Human Reconstruction.}
Beyond articulated pose estimation, markerless MoCap increasingly targets full 3D reconstruction of dynamic humans. These methods aim to recover temporally consistent 3D representations of human motion, including parametric meshes or volumetric reconstructions, from single \cite{biggs2020multibodies, rong2020frankmocap, joo2018total, goel2023humans, ye2023slahmr} or multi-view observations \cite{newcombe2015dynamicfusion, Joo_2017_TPAMI}. Yet, multi-view RGB-D sensing plays a crucial role in a reliable solution, as depth improves geometric completeness and surface fidelity, particularly in dynamic scenes. Moreover, accurate evaluation of methods requires precisely synchronized sensor streams and accurate ground-truth motion capture. In multi-person environments, even small temporal misalignments between RGB-D sensors can introduce geometric artifacts that significantly degrade reconstruction quality. Many existing datasets either lack hardware-level synchronization across cameras or do not provide professional marker-based MoCap ground truth for complex human interactions. Thus, systematic benchmarking of markerless multi-person reconstruction methods remains limited. The absence of synchronized RGB-D observations aligned with robust MoCap supervision poses a barrier to advancing 3D human reconstruction under realistic interaction conditions.

% \subsubsection*{Discussion.} Collectively, prior work demonstrates the critical role of datasets in shaping markerless MoCap development. While existing benchmarks have enabled significant progress in 2D pose, 3D joint estimation, and parametric mesh recovery, they remain limited in their ability to represent realistic, highly interactive multi-person motion. The absence of synchronized multi-view RGB-D data, combined with professional marker-based supervision under complex occlusion and interaction scenarios, leaves a measurable gap between benchmark performance and real-world encounters. The proposed HUM4D is designed specifically to address this gap and to provide a challenging, unified evaluation platform for next-generation markerless MoCap systems in human modeling.

% And it is expected 

% improved performance with more varied actions in our dataset to handle wider pose variation

% However, current datasets cannot represent complex and realistic scenes. We would see improved performance with more varied actions in our dataset to handle wider pose variation.

%great predictions on pose with available datasets such as 
%great predictions on pose with available datasets such as 
\section{HUM4D: Proposed Dataset}\label{sec:hum4d dataset}
% Our survey reveals the critical role of datasets in shaping markerless MoCap development, and there is a measurable gap between benchmark performance and real-world encounters. The proposed HUM4D is designed specifically to address this gap and to provide a challenging, unified evaluation platform for advancing markerless MoCap systems in human modeling.

The proposed HUM4D is designed to provide a challenging, unified evaluation platform for advancing markerless MoCap systems in human modeling.

% While existing benchmarks have enabled significant progress in 2D pose, 3D joint estimation, and parametric mesh recovery, they remain limited in their ability to represent realistic, highly interactive multi-person motion. The absence of synchronized multi-view RGB-D data, combined with professional marker-based supervision under complex occlusion and interaction scenarios, leaves a measurable gap between benchmark performance and real-world encounters. The proposed HUM4D is designed specifically to address this gap and to provide a challenging, unified evaluation platform for next-generation markerless MoCap systems in human modeling.

\begin{figure}[t]
    \centering
\includegraphics[width=\linewidth]{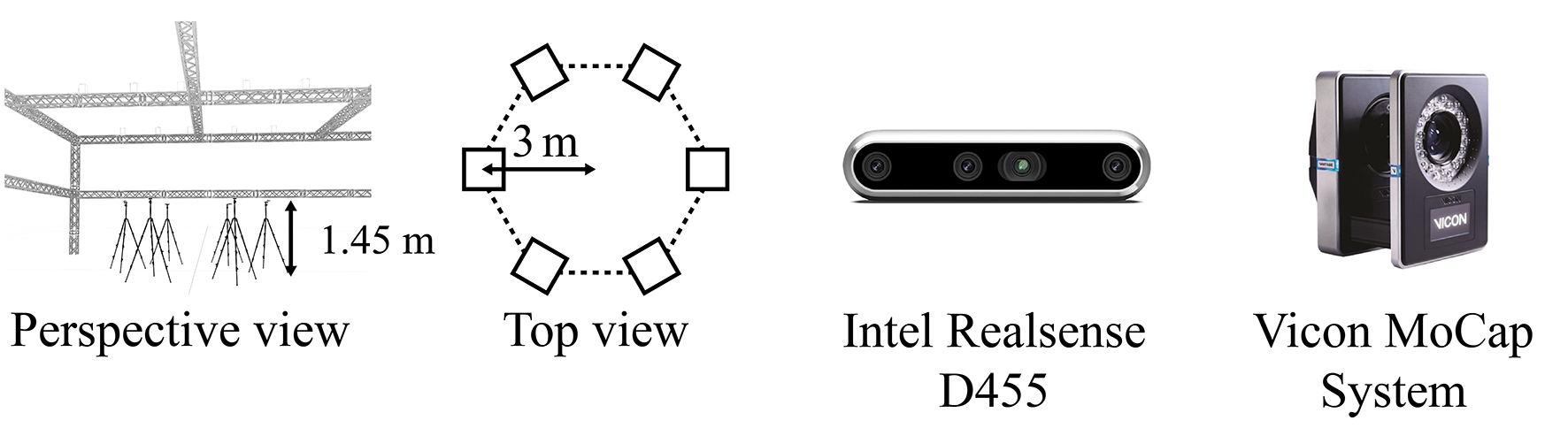}
\caption{\textbf{Hardware setup.} Data acquisition configuration used in HUM4D. From left to right: perspective view of the RGB-D camera placement at approximately 1.45m height; top-view layout showing six cameras arranged in a circular configuration with a 3m radius; Intel RealSense D455 RGB-D sensor used for color and depth capture; and the Vicon motion capture system used for marker-based ground-truth acquisition.}\label{fig:hardwaresetup-fig}
\end{figure} 

\begin{figure*}[t]
    \centering
\includegraphics[width=\linewidth]{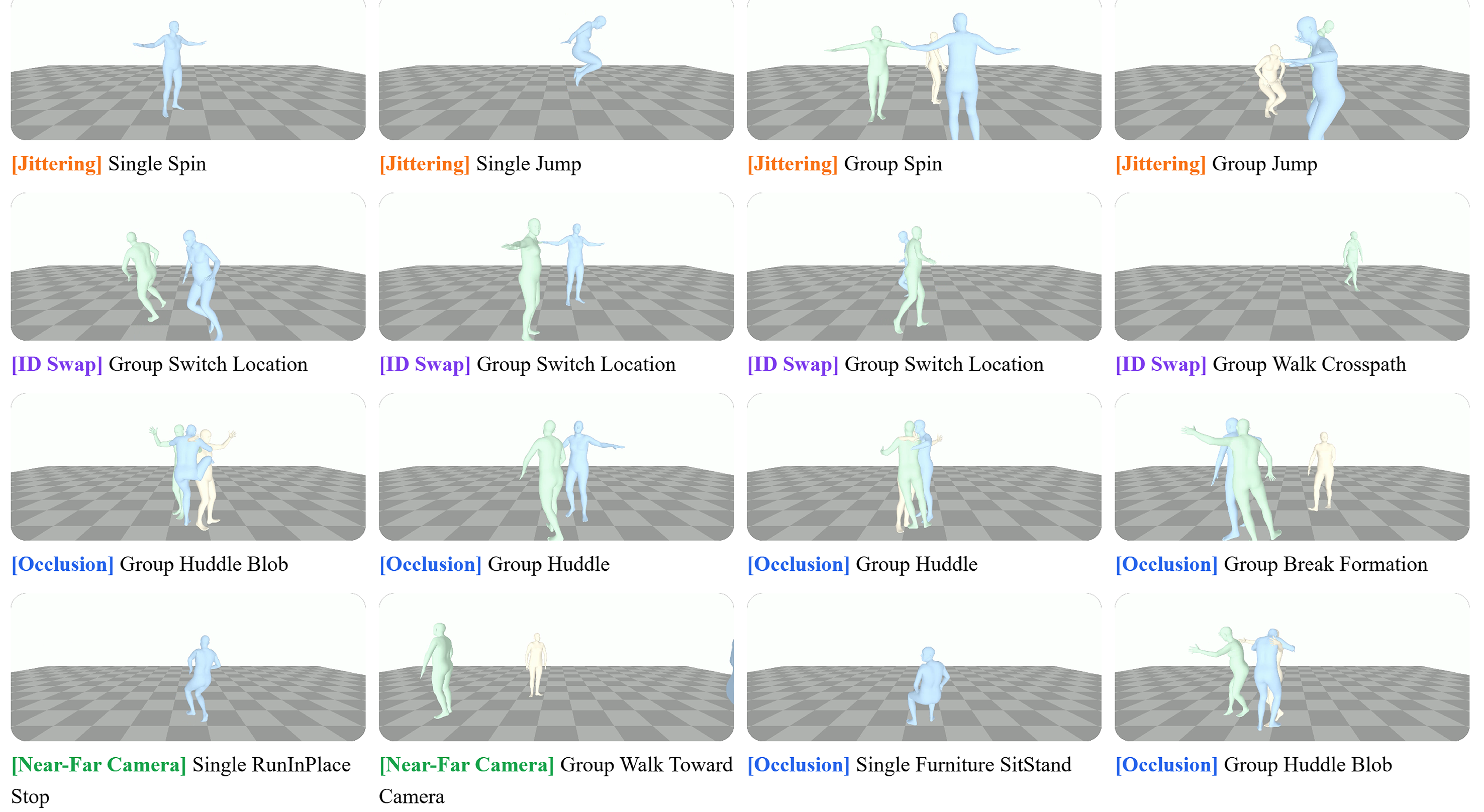}
\caption{3D acquisition of complex human motion sequences captured under challenging conditions such as Jittering, ID Swap, Occlusion, and Near-Far Camera.}\label{fig:different-activities}
\end{figure*}

\smallskip
\formattedparagraph{\textit{(i)} Capture Environment and Hardware Setup.}
The proposed HUM4D dataset is captured in a professional MoCap studio equipped with a $149 ~\textrm{m}^2$ Vicon MoCap stage. The studio contains $44$ synchronized Vicon infrared cameras mounted on a truss system. The installed system enables full-body marker tracking within a large calibrated capture volume. The Vicon system recorded 3D skeleton motion and reflective marker trajectories at $120 ~\textrm{fps}$. 

Participants wore professional MoCap suits equipped with $56$ reflective markers to enable full-body tracking. We calibrated the Vicon system using their proprietary calibration pipeline to ensure sub-millimeter spatial accuracy. RGB-D intrinsic parameter calibration is performed using a checkerboard-based calibration procedure provided by the Intel RealSense SDK. In contrast, temporal synchronization across RGB-D streams is achieved via hardware-level synchronization to ensure consistent multi-view alignment. 

For dense and detailed multi-view 4D acquisition emulating real noise levels, we deployed $6$ Intel RealSense D-455 cameras to capture RGB and depth data. The cameras were mounted on tripods in a 360$^\circ$ configuration surrounding the capture area, forming a circular arrangement with an approximate $6\textrm{m}$ diameter. Each camera was placed $1.45\textrm{m}$ above ground level to approximate human eye level and maximize coverage of upper and full-body motion. The capture environment and hardware setup configuration are shown in Fig. \ref{fig:capture-stage} and Fig. \ref{fig:hardwaresetup-fig}, respectively.

We chose the aforementioned capture space volume to ensure that all participants remain within the field of view of the RGB-D sensors throughout the performance sequences. Here, the selection of the D455 sensor for capture is primarily due to its widespread use in robotics and computer vision research for its reliable active depth sensing and global-shutter RGB imaging. We operated all cameras at $720$p resolution to balance spatial coverage, depth accuracy, and storage requirements. Although the D455 supports depth sensing up to approximately $6\textrm{m}$, its optimal accuracy is achieved within $3\textrm{m}$. Therefore, camera placement and activity design were constrained and adjusted accordingly.

\smallskip
\formattedparagraph{\textit{(ii)} 4D Acquisition.}
% Three male actors participated in the data collection. The dataset comprises multimodal recordings of 52 distinct action sequences, including 14 single-person actions and 41 multi-person actions interactions. The total dataset contains 83,768 frames.
%
% The recorded activities span physical exercises, daily activities, and socially interactive behaviors. Importantly, the multi-person sequences were designed to induce challenging motion capture conditions, including frequent inter-person occlusions, close physical contact, identity swaps, and motion blur. Several scenarios involve tightly coupled three-person group interactions, significantly increasing reconstruction ambiguity and volumetric overlap.
%
% During acquisition, actors wore MoCap suits with 56 reflective markers to ensure accurate skeletal tracking. The MoCap system recorded joint positions and orientations at 120\,fps. For highly occluded scenarios (e.g., hugging or clustered formations), additional unlabeled marker clusters were attached to preserve tracking continuity and reduce marker dropout.
%
% Table~\ref{table:human4d} summarizes representative single-person and multi-person activities included in HUM4D. The dataset explicitly includes motion types such as jittering motion, severe occlusion, near-far camera interaction, and identity switching, enabling systematic evaluation of multi-person markerless motion capture under realistic and challenging conditions.
%
To ensure high-fidelity 4D human motion capture, we employed a professional marker-based motion capture system synchronized with multi-view visual sensors. During acquisition, actors wore MoCap suits equipped with 56 reflective markers, enabling accurate skeletal tracking of joint positions and orientations. The system recorded motion trajectories at $120 ~\textrm{fps}$, providing temporally dense and geometrically precise ground truth. 

Capturing tightly coupled multi-person activities and interactions presents tracking challenges due to severe occlusions and marker dropout. To mitigate these effects, we attached additional unlabeled marker clusters in highly occluded configurations, such as hugging, clustered formations, and close-contact interactions. This design choice greatly improved tracking continuity and preserved identity consistency in densely interactive scenarios. The resulting motion data is temporally aligned with synchronized RGB and RGB-D streams, ensuring precise correspondence between visual observations and marker-based ground truth.

The acquisition protocol was intentionally designed to induce challenging reconstruction conditions representative of real-world scenarios. The recorded activities span structured physical exercises, daily functional actions, and socially interactive behaviors. Multi-person sequences include frequent inter-person occlusions, close physical contact, rapid position exchanges between similarly dressed participants, motion blur from fast dynamics, and near-far depth variations relative to the cameras. Several sequences involve tightly coupled three-person group formations, leading to substantial volumetric overlap and reconstruction ambiguity---\textit{conditions that remain underrepresented in existing benchmarks}.

The dataset was collected using three male actors and comprises $52$ distinct action sequences, including $14$ single-person actions and $41$ multi-person interaction sequences. The dataset contains $83,768$ synchronized frames in total. Table~\ref{table:human4d} summarizes representative activities across both single- and multi-person categories with activity type, total number of frames, and motion type. Figure \ref{fig:different-activities} provides a visualization of a few example motions. By combining synchronized multi-view RGB-D observations with professional marker-based ground truth under complex multi-person dynamics, the proposed dataset provides a realistic and demanding testbed for evaluating next-generation markerless 4D human motion capture systems. \textit{Refer to the supplementary for details on activity and motion types.}

\begin{figure*}[t]
    \centering
\includegraphics[width=0.96\linewidth]{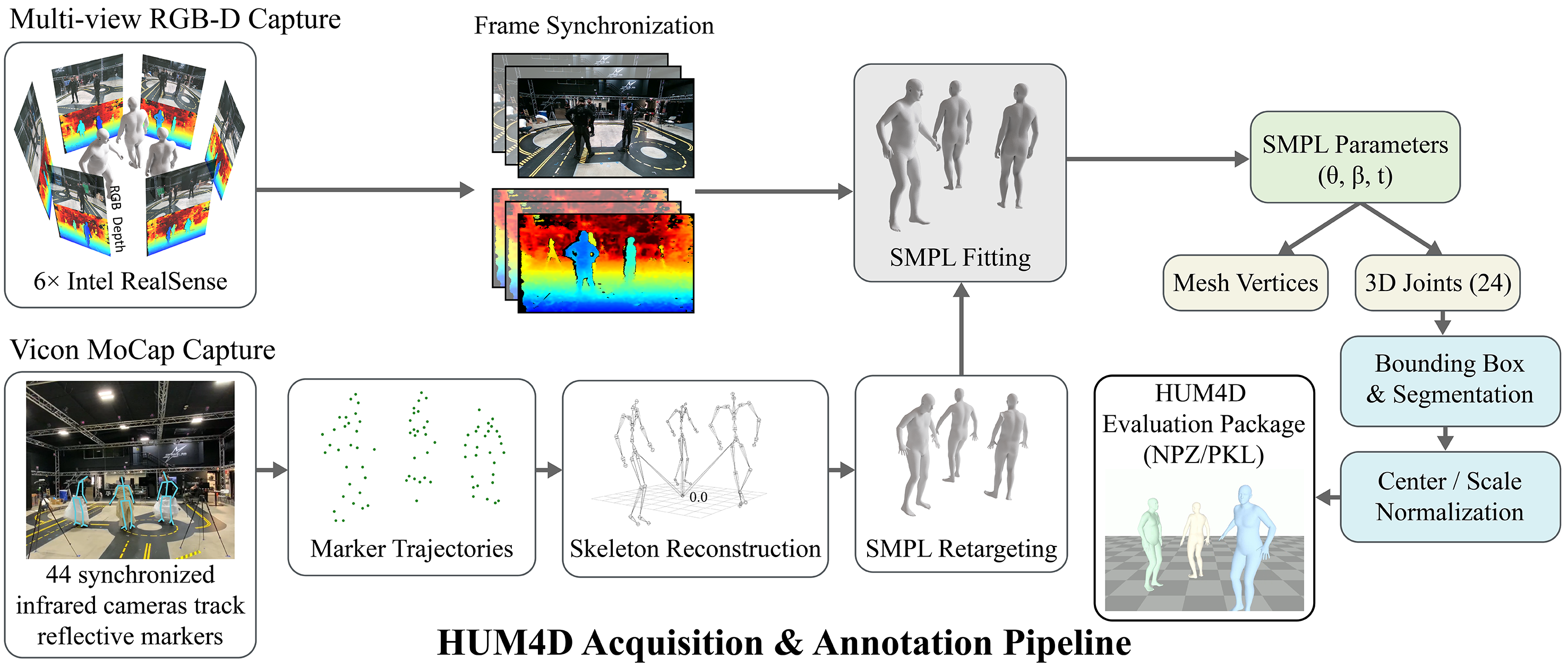}
\caption{Overall acquisition pipeline: six synchronized multi-view RGB-D camera with Vicon MoCap system. RGB, RGB-D, and Vicon frame rate synchronization with temporal alignment is then later followed by processing and annotation to acquire accurate 3D data.}\label{fig:pipeline}
\end{figure*} 

\smallskip
\formattedparagraph{\textit{(iii)} Processing and Annotation.} 
Raw multimodal recordings alone are insufficient for precise benchmarking. To enable reproducible evaluation and direct compatibility with recent methods, the captured data is carefully re-targeted, temporally aligned, and standardized into commonly used parametric formats. Our post-processing and annotation ensures geometric consistency between marker-based ground truth and visual observations. It further provides SMPL-compatible annotations suitable for training and evaluation.

\begin{itemize}
\item \textbf{MoCap export and SMPL retargeting}. All sequences are recorded using the Vicon system at $120 ~\textrm{fps}$ and exported as FBX files containing the skeletal animation and marker-driven motion. Because the native Vicon skeleton differs from the SMPL kinematic structure in joint hierarchy, rest pose, and naming conventions, direct use of the captured motion is not possible. We therefore import each FBX sequence into Maya and perform IK-based retargeting from the Vicon skeleton to the canonical SMPL $24$-joint kinematic tree. This step resolves structural discrepancies and produces a temporally consistent set of SMPL-aligned joint transforms across all sequences.

\item \textbf{FBX to PKL conversion}. Following retargeting, we export SMPL motion parameters using an automated Maya batch-processing script. For each sequence, the script identifies the skinned SMPL mesh and its associated skinCluster, maps mesh influences to the canonical SMPL joints, and iterates over the valid frame range to extract per-joint rotations. Each processed sequence is stored as a PKL file containing: \texttt{pose} ($24$-joint axis-angle rotations per frame), \texttt{trans} (root translation in meters), \texttt{betas} (sequence-level shape coefficients), ordered \texttt{joint names}, and metadata including frame boundaries and frame rate. This representation follows widely adopted conventions in parametric human modeling.

\item \textbf{Temporal alignment and downsampling}. The RGB-D streams are recorded at $15 ~\textrm{fps}$, whereas the Vicon system captures motion at $120 ~\textrm{fps}$. To ensure precise frame-level correspondence between visual observations and motion parameters, we downsample the MoCap data to $15 ~\textrm{fps}$ after retargeting by selecting frames at a fixed stride $(8:1)$. The resulting motion window is exported according to per-sequence frame-range metadata. This approach guarantees one-to-one temporal alignment between the RGB-D frames and the associated SMPL parameters.

\item \textbf{3D keypoints and bounding boxes}. To facilitate standardized benchmarking, we additionally provide per-frame SMPL-aligned 3D joint coordinates and person-centric crop metadata. Specifically, we provide a \texttt{.npz} file containing \texttt{D keypoints} with shape $(\textrm{T}, 24, 3)$ in meters, corresponding image identifiers \texttt{imgname}, and crop parameters including \texttt{bbox\_xyxy}, \texttt{center}, and \texttt{scale}. This unified format supports direct evaluation of single-view and multi-view pose and mesh reconstruction pipelines under consistent preprocessing assumptions.
\end{itemize}

\noindent
Figure \ref{fig:pipeline} shows the overall acquisition pipeline. Refer to the supplementary material for the folder structure and file arrangement within each motion type shown in Table \ref{table:human4d}.

\begin{table}
\centering
\caption{HUM4D physical, daily and social activities.}
\label{table:human4d}
\resizebox{1.0\linewidth}{!}{
\begin{tabular}{lcc}
\hline
\rowcolor{gray!20}
\textbf{Activity} & \textbf{\# Frames} & \textbf{Motion Type} \\
\hline
\rowcolor{red!20}
\multicolumn{3}{c}{\textbf{Single-person}} \\
\hline
Single\_Spin & 3,653 & Jittering \\
Single\_Jump & 4,603 & Jittering \\
Single\_RunInPlace\_Stop & 3,582 & Jittering \\
Single\_Furniture\_SitStand & 7,660 & Occlusion \\
\hline
\rowcolor{blue!20}
\multicolumn{3}{c}{\textbf{Multi-person}} \\
\hline
Group\_Spin & 2,388 & Jittering \\
Group\_Jump & 3,291 & Jittering \\
Group\_Walk\_CrossPath & 6,903 & Jittering \\
Group\_Huddle & 6,586 & Occlusion \\
Group\_Huddle\_Blob & 11,402 & Occlusion \\
Group\_Break\_Formation & 6,783 & Occlusion \\
Group\_Walk\_Toward\_Camera & 10,155 & Near Far Camera \\
Group\_Run\_Around & 6,141 & ID Swap \\
Group\_Switch\_Location & 7,962 & ID Swap \\
Group\_Hide\_Each\_Other & 2,659 & ID Swap \\
\hline
\textbf{Total} & \textbf{83,768} & \\
\hline
\end{tabular}
}
\end{table}

\section{Evaluation and Benchmarking}
\label{sec:benchmark}
We evaluate the utility of the proposed HUM4D by benchmarking state-of-the-art human mesh recovery methods under a cross-dataset generalization setting. All models are evaluated using publicly available weights without fine-tuning on HUM4D. This protocol isolates the dataset's generalization challenge and prevents in-domain adaptation from masking inherent difficulty.

%\subsection{Evaluation}

% \textcolor{red}{Refine from here. The chronology should be 1) About our data 2) Evaluation}

\subsection{Evaluation}
The proposed HUM4D is a multi-person human 4D motion dataset designed to stress-test 3D human reconstruction under close interactions, severe interaction occlusions, and frequent body overlap. It contains $52$ motion capture sequences aligned with synchronized RGB image streams. For benchmarking, we use all available aligned frames, resulting in $83,768$ RGB frames temporally aligned with SMPL-based 3D ground-truth joints derived from a professional marker-based motion capture system.

\smallskip
\noindent
\textbf{Ground-truth generation.}
Ground-truth 3D joints are generated by forwarding the SMPL model using exported pose (in axis-angle representation), global translation (in meters), and sequence-level shape parameters. This yields consistent and physically plausible joint locations while preventing the fidelity of the underlying motion capture. The resulting SMPL joints serve as the reference for evaluating pose reconstruction quality across all methods.

\smallskip
\noindent
\textbf{Temporal alignment and synchronization.}
RGB frames and motion capture frames are synchronized at $15 ~\textrm{fps}$. When image sequences are shorter than motion capture recordings, sequences are truncated to ensure strict frame-level correspondence and synchronization. This protocol avoids temporal leakage and ensures that errors reflect reconstruction difficulty rather than synchronization artifacts.

\smallskip
\formattedparagraph{Evaluation Metric.} Following prior work in this area \cite{Hao_2025_ICCV}, we report Procrustes-Aligned Mean Per Joint Position Error (PA-MPJPE) in millimeters. Given predicted 3D joints $\hat{J}$ and ground-truth joints $J$, PA-MPJPE is computed after rigid alignment (scale, rotation, and translation) via Procrustes analysis \cite{gower2004procrustes}

\begin{equation}
\text{PA-MPJPE} = \frac{1}{N} \sum_{i=1}^{N} \left\| J_i - \hat{J}_i^{\text{PA}} \right\|_2,
\end{equation}
where $\hat{J}^{\text{PA}}$ denotes the aligned prediction and $N$ is the number of points. PA-MPJPE isolates pose reconstruction quality independent of global scale and camera coordinate inconsistencies. This makes it suitable for cross-dataset evaluation where intrinsic calibration and absolute scale assumptions may differ between training and test domains.

% We use PA-MPJPE because it isolates pose reconstruction quality independent of global scale and camera coordinate inconsistencies. This makes it suitable for cross-dataset evaluation where intrinsic calibration and absolute scale assumptions may differ between training and test domains.

 \subsection{Representative Baselines and Observations}
We evaluate four representative frame-based methods for benchmarking. These include \textbf{SPIN} \cite{kolotouros2019spin}, \textbf{PARE} \cite{Kocabas_PARE_2021}, \textbf{HMR2.0} \cite{goel2023humans}, and \textbf{PersPose} \cite{Hao_2025_ICCV}. These methods are selected 
to cover approaches with modeling diversity, strong baseline performance, and controlled evaluation conditions that cover both traditional and deep-learning approaches to human 4D MoCap. Our goal is to assess generalization behavior across distinct modeling strategies rather than to favor a particular design paradigm. All selected methods report competitive PA-MPJPE on 3DPW \cite{vonMarcard2018}, ensuring we benchmark against robust, widely adopted baselines.

Table \ref{table:hom4d_benchmark} summarizes the quantitative evaluation of representative methods on our HUM4D dataset compared to their reported performance on the popular 3DPW \cite{vonMarcard2018} benchmark. While all evaluated approaches achieve competitive accuracy on 3DPW, their performance deteriorates drastically when tested on HUM4D. PA-MPJPE increases from approximately $39–82 ~\textrm{mm}$ on 3DPW to $151–180 ~\textrm{mm}$ on HUM4D, indicating a severe cross-dataset generalization gap even for methods as recent at year 2025. This clearly indicates the need for a new dataset covering  
dense multi-person interactions, severe occlusions, and frequent identity exchanges for developing better markerless MoCap model.

% This performance gap highlights the increased difficulty and challenges posed by HUM4D, which includes dense multi-person interactions, severe occlusions, and frequent identity exchanges. 

\subsubsection{Specific Observations.}

\formattedparagraph{A. Impact of Interaction-Induced Occlusion.}
One of the dominant challenges present in HUM4D arises from persistent inter-person occlusion and volumetric overlap. Many sequences involve tightly coupled interactions, such as group huddles, crossing trajectories, and close-contact motion, during which large portions of the body are temporarily invisible. Methods such as SPIN \cite{kolotouros2019spin} and HMR2.0 \cite{goel2023humans} rely heavily on reliable 2D keypoint evidence extracted from single images. When limbs overlap or become occluded by another person, keypoint detectors produce ambiguous or missing observations, which propagate errors to the downstream SMPL \cite{loper2023smpl, loper2015smpl} regression. As a result, optimization-based refinement strategies, such as those used in SPIN \cite{kolotouros2019spin}, often fail to recover plausible body configurations when the initial 2D evidence is corrupted.

\smallskip
\formattedparagraph{B. Robustness of Part-Aware and Geometry-Aware Models.} Among all the evaluated methods, PersPose \cite{Hao_2025_ICCV} achieves the lowest error on HUM4D. This result suggests that explicitly modeling camera perspective and geometric constraints can partially mitigate ambiguities introduced by close-range interactions and depth variation. In HUM4D, subjects frequently move toward or away from the cameras, creating significant depth disparities between body parts. Perspective-aware modeling helps reduce these ambiguities by enforcing geometric consistency during pose reconstruction. PARE \cite{Kocabas_PARE_2021}, which emphasizes part-level attention, demonstrates moderate robustness under partial occlusion. By focusing on visible body regions and down-weighting occluded areas, PARE can maintain stable predictions for isolated limbs. Yet, HUM4D contains many scenarios in which even the torso and pelvis, often the most informative regions, are occluded by body overlap. Under such conditions, the advantages of part-based reasoning diminish.

% Evaluation metrics include PA-MPJPE on 3DPW and Occ-MPJPE on HUM4D-Hard. All the evaluation presented in this table is conduced on PA-MPJPE.

\begin{figure}[t]
    \centering
\includegraphics[width=\linewidth]{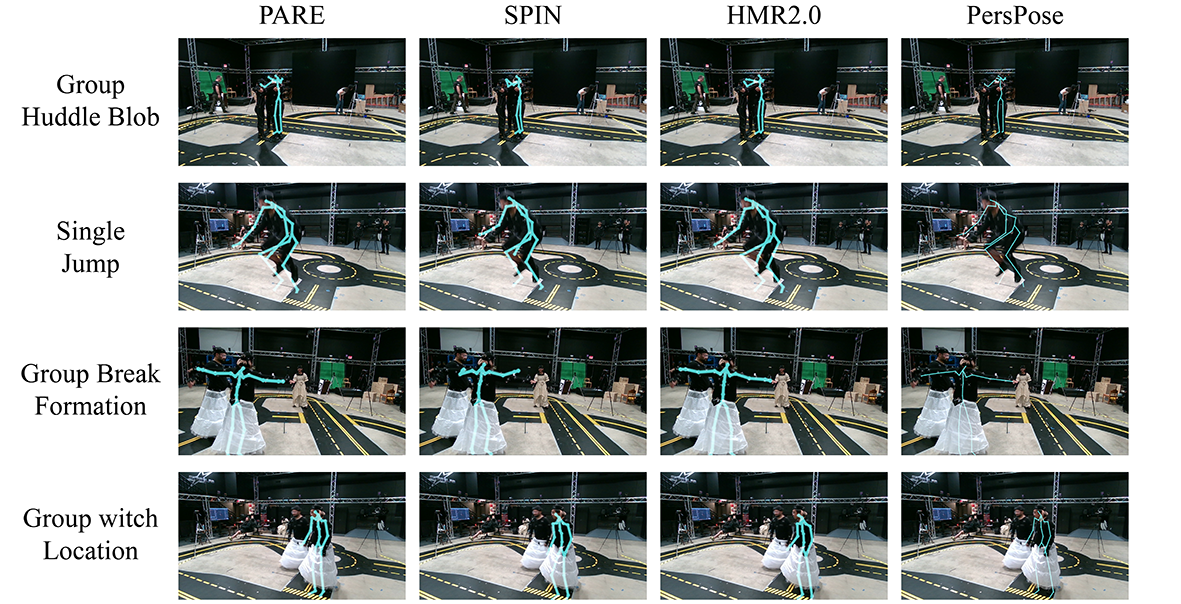}
\caption{\textbf{Human mesh recovery on HUM4D.} Predicted SMPL meshes from SPIN \cite{kolotouros2019spin}, PARE \cite{Kocabas_PARE_2021}, HMR2.0 \cite{goel2023humans}, and PersPose \cite{Hao_2025_ICCV} are overlaid on a challenging frame.}\label{fig:capture-env}
\end{figure}

\smallskip
\formattedparagraph{C. Limitations of Frame-Based Reconstruction.} Another key observation is that all evaluated approaches operate on single frames without explicitly modeling multi-person interactions or temporal consistency. In densely interactive scenes, multiple plausible body configurations can explain the same image evidence, particularly when bodies overlap or exchange positions. Without temporal or interaction-aware constraints, frame-based methods fail to maintain stable identity assignments and coherent pose trajectories.

Table \ref{table:hom4d_benchmark} results reveal that current markerless MoCap models remain heavily optimized for datasets containing single-person actions and moderate occlusions. HUM4D exposes limitations of these models such as: 1) Lack of interaction-aware modeling for multi-person scenes, 2) Insufficient robustness to severe occlusion and body overlap, and 3) Limited exploitation of multi-view and temporal cues. These observations suggest that future markerless MoCap systems must use more than single-frame data and incorporate interaction reasoning, temporal modeling, and multi-view geometric information to achieve reliable performance in realistic environments. Overall, the observation demonstrates that HUM4D constitutes a substantially more challenging benchmark than existing datasets and provides a valuable testbed for developing future markerless multi-person 4D human MoCap.

\begin{table}[t]
%\resizebox{1.0\linewidth}{!}{
\begin{tabular}{l|cccc}
\hline
\textbf{Frame Based Methods} & \textbf{3DPW}\cite{vonMarcard2018} & \textbf{HUM4D}\\
\hline
PARE ~\cite{Kocabas_PARE_2021} & 82.0 & \textbf{177.2}\\
SPIN ~\cite{kolotouros2019spin} & 59.2 & \textbf{179.1} \\
HMR2.0 a~\cite{goel2023humans} & 81.3 & \textbf{180.0}\\
PersPose ~\cite{Hao_2025_ICCV} & 39.1 &  \textbf{151.9}\\
\hline
\end{tabular}
%}
\caption{Baseline PA-MPJPE result on HUM4D compared to 3DPW on which these methods were originally trained/tested.}
\label{table:hom4d_benchmark}
\end{table}

\section{Conclusion and Future Extension}
\label{sec:conclusion}
% In this paper, we introduce HUM4D, a multi-view RGB-D dataset with professional marker-based motion capture ground truth, designed to capture complex multi-person interactions. Through benchmarking state-of-the-art human mesh recovery methods, we observe substantial performance degradation compared to standard datasets such as 3DPW, highlighting the limitations of current approaches under severe occlusion and identity switching. Extensive evaluation demonstrates the increased difficulty of HUM4D and underscores its potential as a realistic and challenging benchmark for advancing multi-person 3D human reconstruction. These results motivate future work on interaction-aware reconstruction, stronger occlusion reasoning, and the incorporation of temporal and multi-person constraints beyond single-frame regression.

We introduced HUM4D, a multi-view RGB-D dataset with professional marker-based motion capture ground truth for complex multi-person markerless human motion capture. Unlike existing benchmarks that largely emphasize single-person or moderately interactive scenarios, HUM4D captures challenging real-world conditions, including severe inter-person occlusions, close-contact interactions, identity switching, and significant depth variation. The dataset provides synchronized RGB-D streams, precise calibration, and temporally aligned SMPL annotations, enabling reliable evaluation of modern human mesh recovery methods.
Benchmarking representative state-of-the-art approaches reveals a substantial generalization gap, i.e., models that achieve strong performance on standard datasets degrade significantly on HUM4D. These results highlight fundamental limitations of current frame-based reconstruction pipelines under dense interactions and heavy occlusion. HUM4D opens several avenues for future research, including interaction-aware reconstruction, temporal reasoning for identity-consistent motion modeling, and multi-view geometry-aware learning that better exploits RGB-D observations. We hope the dataset will serve as a challenging benchmark for developing robust markerless MoCap systems that operate reliably in multi-person environments.

\formattedparagraph{Acknowledgment.} The authors thank Michael Walsh and Morgan Jenks for helping us with the human motion capture acquisition at the RELLIS Starlab facility at Texas A\&M University (TAMU). We also acknowledge the valuable discussions and feedback from John Keyser of the Department of CSCE at TAMU. Additionally, we thank Jyothi Naidu for the support in facilitating the IRB approval process. Finally, the authors thank TAMU's High Performance Research Computing (HPRC) facility for providing startup credits for using the server GPUs.

{
    \small
    \bibliographystyle{ieeenat_fullname}
    \bibliography{main}
}

\clearpage
\setcounter{page}{1}
\maketitlesupplementary
%\title{A Dataset and Evaluation for Complex 4D Markerless Human Motion Capture ---Supplementary Material---}

\begin{abstract}
    %Write an abstract that reads like a contiuation of the main paper. Example: 
    Continuing with our main paper, this supplementary material provides additional details on the motion activities present in HUM4D and how the dataset is organized. First, we describe motion activities present in the dataset and how it is captured, with visual examples. We then present the folder structure of the proposed dataset with a flow chart for easier understanding.
\end{abstract}

\section{Motion and Activity Type}

\begin{figure}
    \centering
\includegraphics[width=\linewidth]{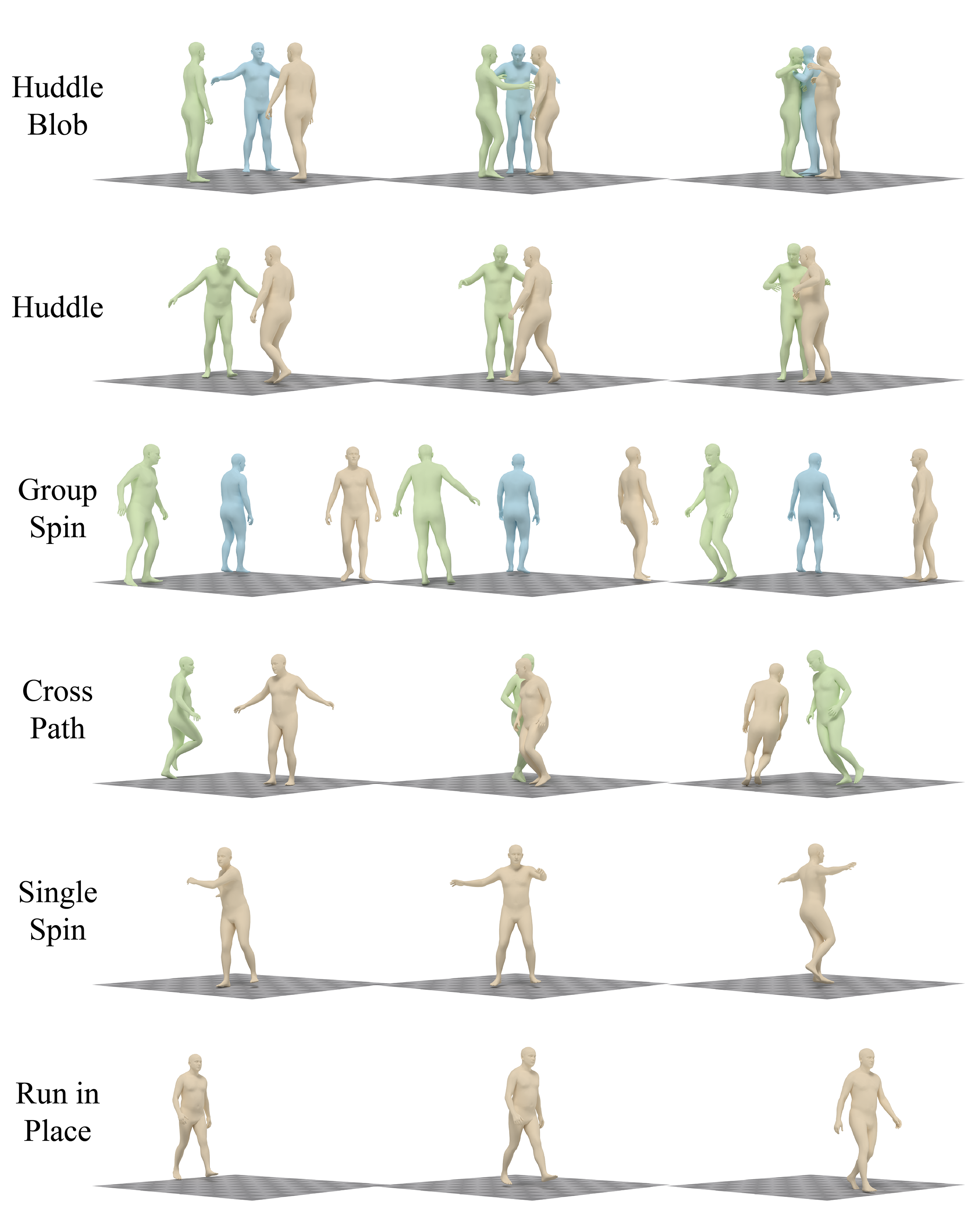}
\caption{Representative examples of motion activities in HUM4D grouped by motion type.}\label{fig:motion_examples}
\end{figure}

In this supplementary, we provide a more detailed description of the motion categories in HUM4D and explain how the dataset is organized. HUM4D is designed to capture challenging motion patterns that are not sufficiently represented in existing markerless motion-capture benchmarks, including rapid local motion, heavy interaction occlusion, identity ambiguity, and depth variation. The dataset groups activities into four motion types, namely \textbf{Jittering}, \textbf{Occlusion}, \textbf{Near Far Camera}, and \textbf{ID Swap}. Representative examples for each motion type are provided in Fig.~\ref{fig:motion_examples}.

\begin{figure*}
    \centering
\includegraphics[width=\linewidth]{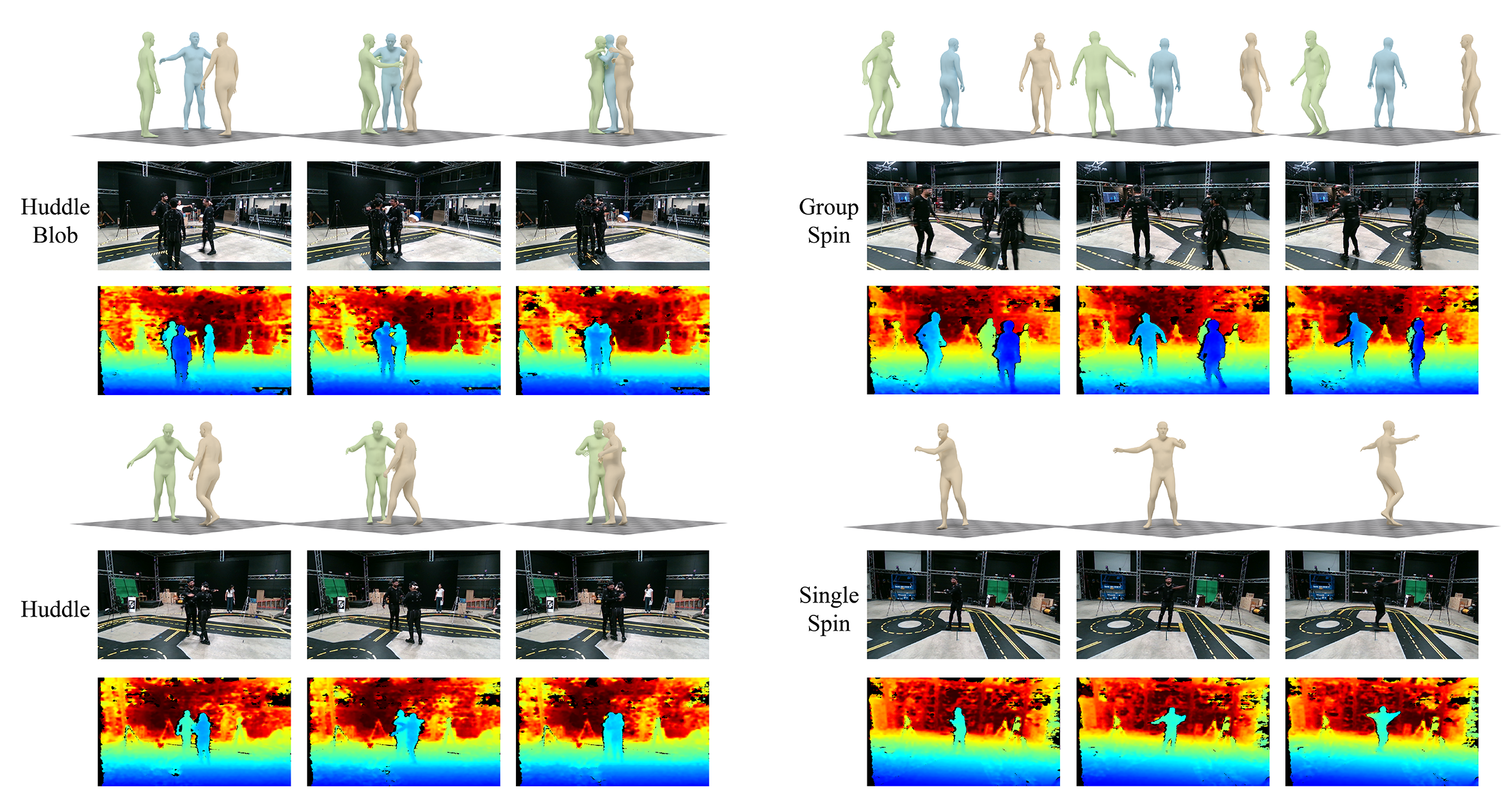}
\caption{Representative examples from HUM4D. Each example includes the body-model or MoCap visualization, the synchronized RGB image, and the corresponding depth map.}\label{fig:image_examples}
\end{figure*} 

\smallskip
\noindent
\textbf{\textit{(i)} Jittering} refers to motion sequences with rapid or highly dynamic body movements that are difficult to estimate consistently across time. This category is intended to stress-test temporal stability and robustness under fast articulation changes, sudden pose transitions, and rapid appearance changes caused by motion.

\begin{itemize}
\item \textbf{Single Spin}: A single subject continuously rotates the body, producing fast orientation changes.
\item \textbf{Single Jump}: A single subject performs repeated jump motion with strong vertical displacement and fast pose transitions.
\item \textbf{Single Run in Place}: A single subject performs running motion in place, often followed by sudden stopping, which introduces rapid temporal changes in limb dynamics.
\item \textbf{Group Spin}: Multiple participants rotate simultaneously, increasing temporal ambiguity and making consistent tracking more difficult.
\item \textbf{Group Jump}: Multiple subjects perform fast jump motions together.
\item \textbf{Group Cross Path}: Multiple subjects repeatedly walk in crossing directions.
\end{itemize}

Overall, the jittering category is designed to evaluate how well a method handles fast body motion and temporal inconsistency.

\smallskip
\noindent
\textbf{\textit{(ii)} Occlusion.} includes activities in which body parts become partially or heavily hidden due to self-occlusion, interaction overlap, or close formation changes. These sequences are intended to evaluate robustness when visual evidence is incomplete. 

\begin{itemize}
\item \textbf{Single Furniture Sit Stand}: A single subject repeatedly sits on and stands up from furniture, causing partial body occlusion and self occlusion
\item \textbf{Group Huddle}: Multiple participants gather closely together, producing severe interaction overlap and limited visibility of individual body parts.
\item \textbf{Group Huddle Blob}: Multiple subjects form a dense cluster, creating heavy body overlap and strong ambiguity in person body association.
\item \textbf{Group Break Formation}: Multiple participants begin in a compact formation and then separate, resulting in changing visibility, overlapping limbs, and dynamic occlusion patterns.
\end{itemize}

These activities create frequent visibility loss and overlapping limbs that are challenging for both 2D keypoint detection and 3D reconstruction.

\smallskip
\noindent
\textbf{\textit{(iii)} Near Far Camera.} captures situations where subjects move toward or away from the cameras, producing substantial scale and depth variation. This category is designed to evaluate robustness to perspective effects and changing camera-relative distance.

\begin{itemize}
\item \textbf{Group Walk Toward Camera}: Multiple subjects walk toward the camera, creating large depth changes, increasing apparent body scale, and introducing viewpoint dependent variation across time.
\end{itemize}

This motion is challenging because large depth changes affect depth estimation and reconstruction quality.

\smallskip
\noindent
\textbf{\textit{(iv)} ID Swap.} refers to motion situations in which multiple people move in close proximity and exchange relative positions, making identity tracking difficult over time. This category is intended to reveal failures in temporal association and person identity consistency.

\begin{itemize}
\item \textbf{Group Run Around}: Multiple participants run around one another, causing frequent changes in relative position and making person identity association challenging.
\item \textbf{Group Switch Location}: Multiple subjects exchange their spatial locations, explicitly testing whether methods can preserve person identity across motion.
\item \textbf{Group Hide Each Other}: Participants move in ways that partially or fully block another, creating temporary disappearance and reappearance that can lead to identity confusion.
\end{itemize}

These sequences often expose failures in methods that depend on consistent person association across frames.

\section{Dataset Arrangement}

\begin{figure*}[t]
    \centering
    \includegraphics[width=\linewidth]{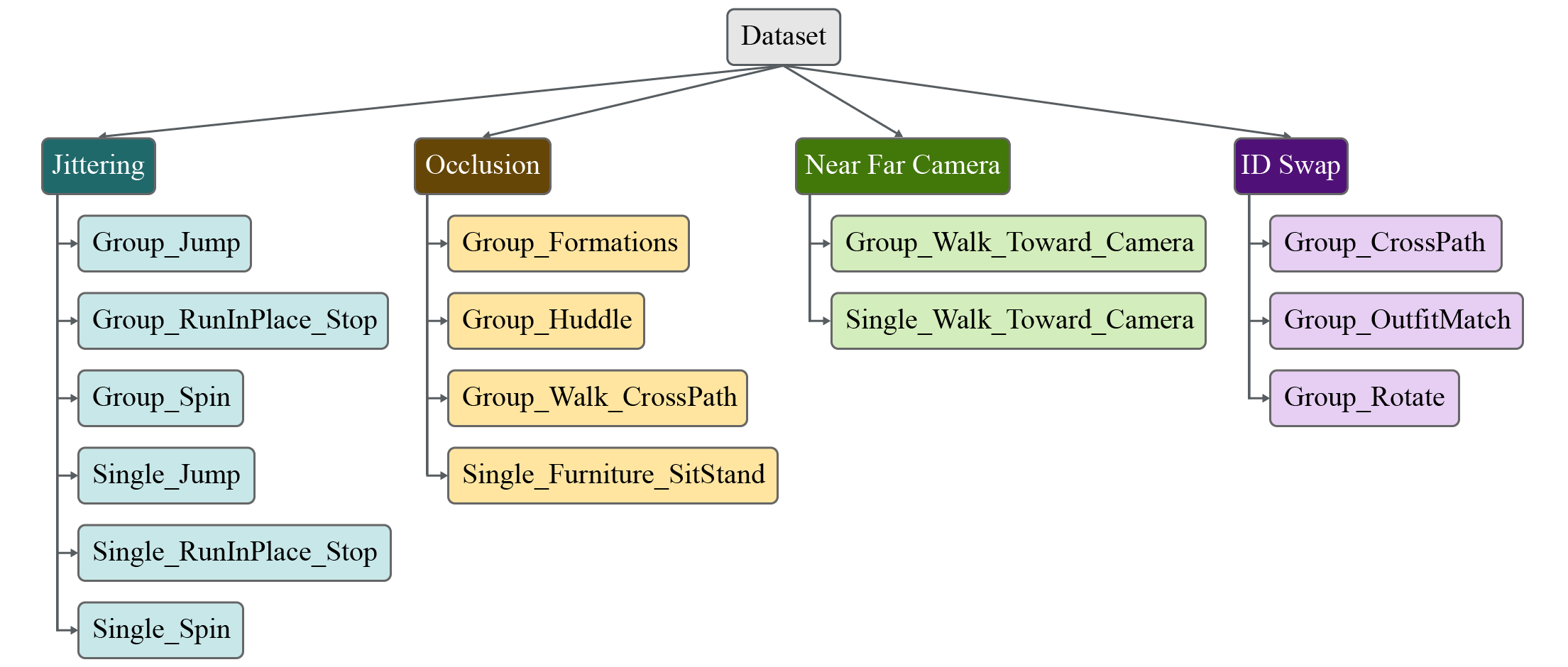}
    \caption{Top level hierarchy of HUM4D. The dataset is first grouped by motion type, and each motion type contains a set of activity folders.}\label{fig:intro-hierarchy1}
\end{figure*}

\begin{figure*}[t]
    \centering
    \includegraphics[width=\linewidth]{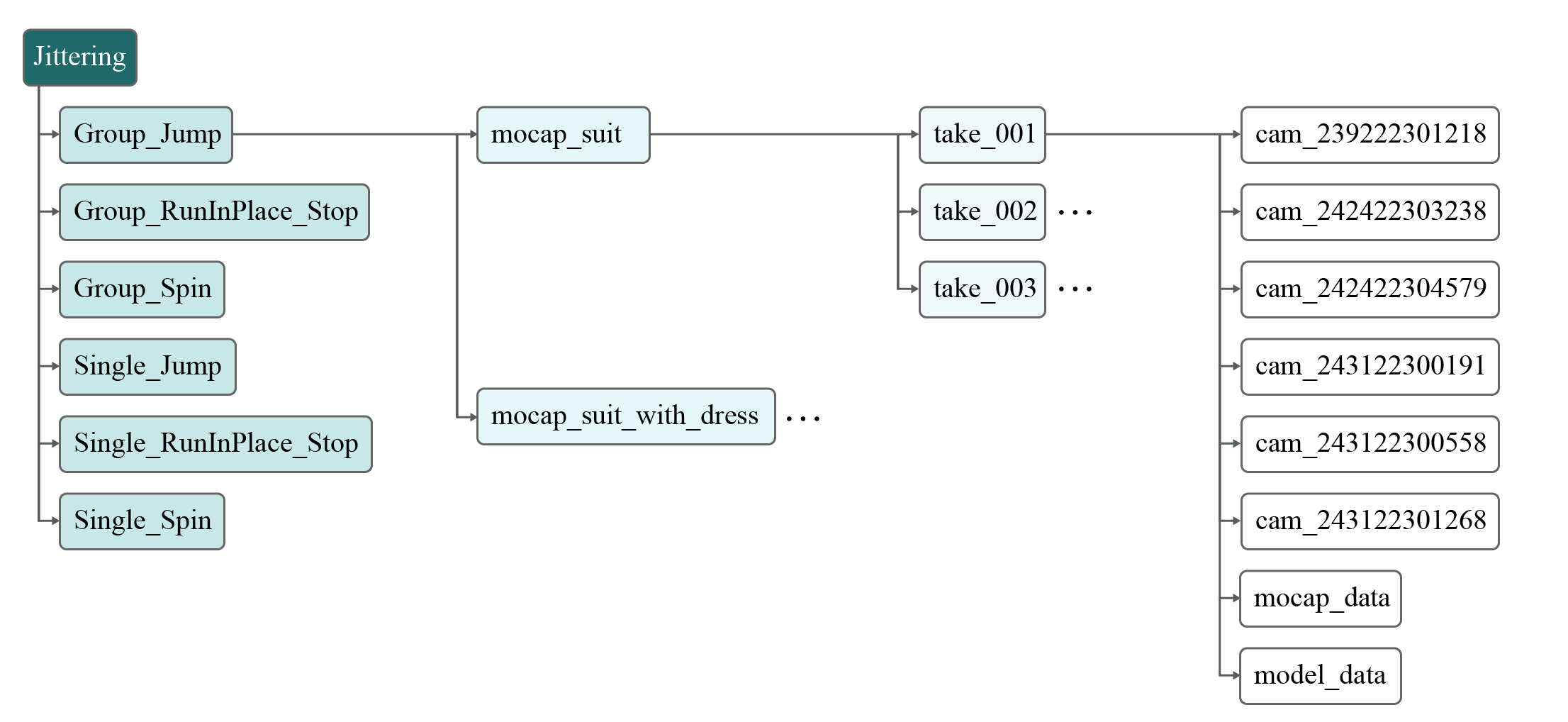}
    \caption{Example lower level hierarchy of HUM4D. Within each activity, the data is further organized by recording setting, take index, multi-view camera streams, and associated annotation files such as mocap data and model data.}\label{fig:intro-hierarchy2}
\end{figure*}

In this section, we describe how HUM4D is organized for convenient access. As illustrated in Fig.~\ref{fig:intro-hierarchy1} and Fig.~\ref{fig:intro-hierarchy2}, the dataset follows a hierarchical structure from motion type to action category, recording setting, take index, and camera streams and annotation files.

At the top level, the dataset is divided into four motion type groups: \textbf{Jittering}, \textbf{Occlusion}, \textbf{Near Far Camera}, and \textbf{ID Swap}. Each of these groups contains multiple activity folders that correspond to the dominant motion pattern shown in that category. For example, the \textbf{Jittering} group contains activities such as \texttt{Group Jump}, \texttt{Group RunInPlace Stop}, \texttt{Group Spin}, \texttt{Single Jump}, \texttt{Single RunInPlace Stop}, and \texttt{Single Spin}. Similarly, the remaining motion groups contain their own activity folders, such as \texttt{Single Furniture Sit Stand}, \texttt{Group Huddle}, \texttt{Group Huddle Blob}, and \texttt{Group Break Formation} under \textbf{Occlusion}, \texttt{Group Walk Toward Camera} under \textbf{Near Far Camera}, and \texttt{Group Run Around}, \texttt{Group Switch Location}, and \texttt{Group Hide Each Other} under \textbf{ID Swap}.

Within each activity folder, the data is further organized by recording setting. This level reflects differences in capture configuration or subject appearance, such as \texttt{mocap suit} and \texttt{mocap suit with dress}. Each recording setting then contains multiple repeated captures indexed by take number. For example, \texttt{take\_001}, \texttt{take\_002}, and \texttt{take\_003}. 

Inside each take, the dataset contains synchronized multi-view camera streams together with motion annotations and processed model outputs. As shown in Fig.~\ref{fig:intro-hierarchy2}, each take includes several camera entries identified by camera-specific names such as \texttt{cam\_239222301218}. In addition to these image streams, each take also contains annotation entries such as \texttt{mocap data}.

\begin{table}[t]
\centering
%\resizebox{1.0\linewidth}{!}{
\begin{tabular}{l|cccc}
\hline
\textbf{Motion Type} & \textbf{PARE} & 
 \textbf{SPIN} & \textbf{HMR2.0} &  \textbf{PersPose} \\
\hline
Jittering        & 177.6 & 175.6 & 181.1 &  197.3\\
Occlusion      & 157.1 & 168.1 & 148.4  & 166.1\\
Near-Far Camera      & 178.6  & 170.6 & 205.7  & 209.9\\
ID Swap   & 265.3 & 268.7 & 260.8  & 267.7 \\
\rowcolor{blue!20}
\textbf{Overall}         & 185.7 & 189.2 & 184.9 & 199.2 \\
\hline
\end{tabular}
%}
\caption{PA-MPJPE (mm) broken down by motion type on HUM4D.}
\label{table:motion_type_analysis}
\end{table}

\section{Motion Type Analysis} To further analyze method behavior on HUM4D, we report a breakdown of reconstruction performance by motion type. Since HUM4D is organized around four challenging motion categories, namely \textbf{Occlusion}, \textbf{ID Swap}, \textbf{Near-Far Camera}, and \textbf{Jittering}, this evaluation offers a more specific view of model behavior. As shown in Table~\ref{table:motion_type_analysis}, \textbf{ID Swap} is the most challenging motion type, producing the highest PA-MPJPE across all evaluated methods. By contrast, \textbf{Occlusion} yields the lowest average error overall, making it the least challenging category among the four in the benchmark. These results suggest that identity preservation under close multi-person interactions remain particularly difficult for existing methods, and they complement the overall benchmark score with more fine-grained insight.

\end{document}